\documentclass[10pt]{article}

\usepackage{listings}
\usepackage{epsfig}
\usepackage{caption,subcaption}
\usepackage{multicol}
\usepackage{amsmath}
\usepackage{url}
\usepackage[a4paper, total={7.1in, 8.5in}]{geometry}

\begin{document}

\newcommand\copyrighttext{ 
\Huge {IEEE Copyright Notice} \\ \\
\large {Copyright (c) 2014 IEEE \\
Personal use of this material is permitted. Permission from IEEE must be obtained for all other uses, in any current or future media, including reprinting/republishing this material for advertising or promotional purposes, creating new collective works, for resale or redistribution to servers or lists, or reuse of any copyrighted component of this work in other works.} \\ \\

{\Large Published in Proceedings of $5^{th}$ International Conference on Advanced Computing \& Communication Technologies (ACCT-2015), Feb. 21-22, 2015, Rohtak, India} \\ \\ 
{DOI: 10.1109/ACCT.2015.114}
}

\twocolumn[
  \begin{@twocolumnfalse}
    \copyrighttext
  \end{@twocolumnfalse}
]

\author{Vipul K. Dabhi \\
\small Department of Information Technology,\\[-0.8ex]
\small Faculty of Technology, Dharmsinh Desai University, Nadiad, Gujarat, INDIA\\
\small \texttt{vipul.k.dabhi@gmail.com}\\
\and
Sanjay Chaudhary\\
\small Institute of Engineering and Technology,\\[-0.8ex]
\small Ahmedabad University, Ahmedabad-380009, Gujarat, INDIA\\
\small \texttt{sanjay.chaudhary@ahduni.edu.in}
}

\title{Developing Postfix-GP Framework for Symbolic Regression Problems}
\date{}

\maketitle

\begin{abstract}
This paper describes Postfix-GP system, postfix notation based Genetic Programming (GP), for solving symbolic regression problems. It presents an object-oriented architecture of Postfix-GP framework. It assists the user in understanding of the implementation details of various components of Postfix-GP. Postfix-GP provides graphical user interface which allows user to configure the experiment, to visualize evolved solutions, to analyze GP run, and to perform out-of-sample predictions. The use of Postfix-GP is demonstrated by solving the benchmark symbolic regression problem. Finally, features of Postfix-GP framework are compared with that of other GP systems. \\

\textbf {\it Keywords-} {Postfix Genetic Programming; Postfix-GP Framework; Object Oriented Design; GP Software Tool; Symbolic Regression}
\end{abstract}

\section{Introduction}
Evolutionary Algorithms (EA) are a group of computational techniques, which employ the theory of natural selection to a population of individuals to generate better individuals. Genetic Programming (GP) is a paradigm of EA which uses hierarchical, tree structure, variable length representation to code solutions of a problem. GP can be used to intelligently search the solution space for finding the optimal solution of a problem.


There are many GP tools (lil-gp, GeneXproTools, GPLab, ECJ, Open BEAGLE) \cite{lilgp,ferreirac,silva,ecj,gagn} developed by GP practitioners. However, none of these address the following demands of end user before applying them to solve symbolic regression problems: (i) ease of use and (ii) small learning curve. Many of these tools are open source and available freely (lil-gp, ECJ, Open BEAGLE, GPLab for MATLAB) \cite{lilgp,ecj,gagn,silva} whereas the rest are available commercially (GeneXproTools) \cite{ferreirac}. Many of these tools necessitate modification in source code in order to generate required experimental environment. Determining the final solution, produced by these tools, demands translation of the output or digging the log files. Due to these reasons, the interest of researchers and engineers in GP may get reduced. Theses reasons motivated us to develop our own GP framework \cite{dabhia2011} which uses the postfix notation for an individual representation.

We have considered the following features which need to be supported by Postfix-GP framework: (i) easy to extend, (ii) simple and quick procedure for the configuration and running of GP, (iii) a set of algorithm implementation for: (a) generating the initial population, (b) selection mechanisms, and (c) genetic operators, (iv) visualization of: (a) Postfix-GP run analysis and (b) evolved solution with statistical measures, (v) one-step and multi-step prediction support, (vi) visualization of results for one-step and multi-step predictions, and (vii) storage and retrieval of evolved solutions to and from file. Postfix-GP has been used in experimental work \cite{dabhia2011}, \cite{dabhib2012}, \cite{tsmapupgp}, \cite{hwpmfrpoaroi}.

This paper presents the design and implementation of Postfix-GP, an object oriented software framework for genetic programming. Section \ref{sec:Introduction to GP} gives introduction to GP. Section \ref{sec:Postfix-GP} presents the design of Postfix-GP. Moreover, the section also gives the implementation details and main features of Postfix-GP. Section \ref{sec:Case Study} presents Postfix-GP as a solution modeling tool by solving the benchmark symbolic regression problem. Section \ref{sec:featurecomparison} compares the features of Postfix-GP with lil-gp \cite{lilgp}, ECJ \cite{ecj}, and JCLEC \cite{jclec} frameworks. This is followed by conclusions in Section \ref{sec:Conclusions}.
 
\section{Introduction to Genetic Programming}
\label{sec:Introduction to GP}

Standard GP \cite{GPOTPOCBMONS} employs a variable length, tree structure scheme for an individual representation. The tree can be used to represent logical expressions (IF-THEN-ELSE), boolean expressions (AND,OR,NOT) or algebraic expressions. The symbolic regression aims to find the functional relationship (mathematical expression) between given instances of inputs-outputs. GP can be used to perform Symbolic Regression (SR). When using GP for solving symbolic regression problems, the user need to specify the following items: (i) GP configuration parameters, (ii) terminal set and function set, (iii) fitness function.  

\begin{figure}[!htbH]
\begin{center}
\includegraphics*[width=3.4in] {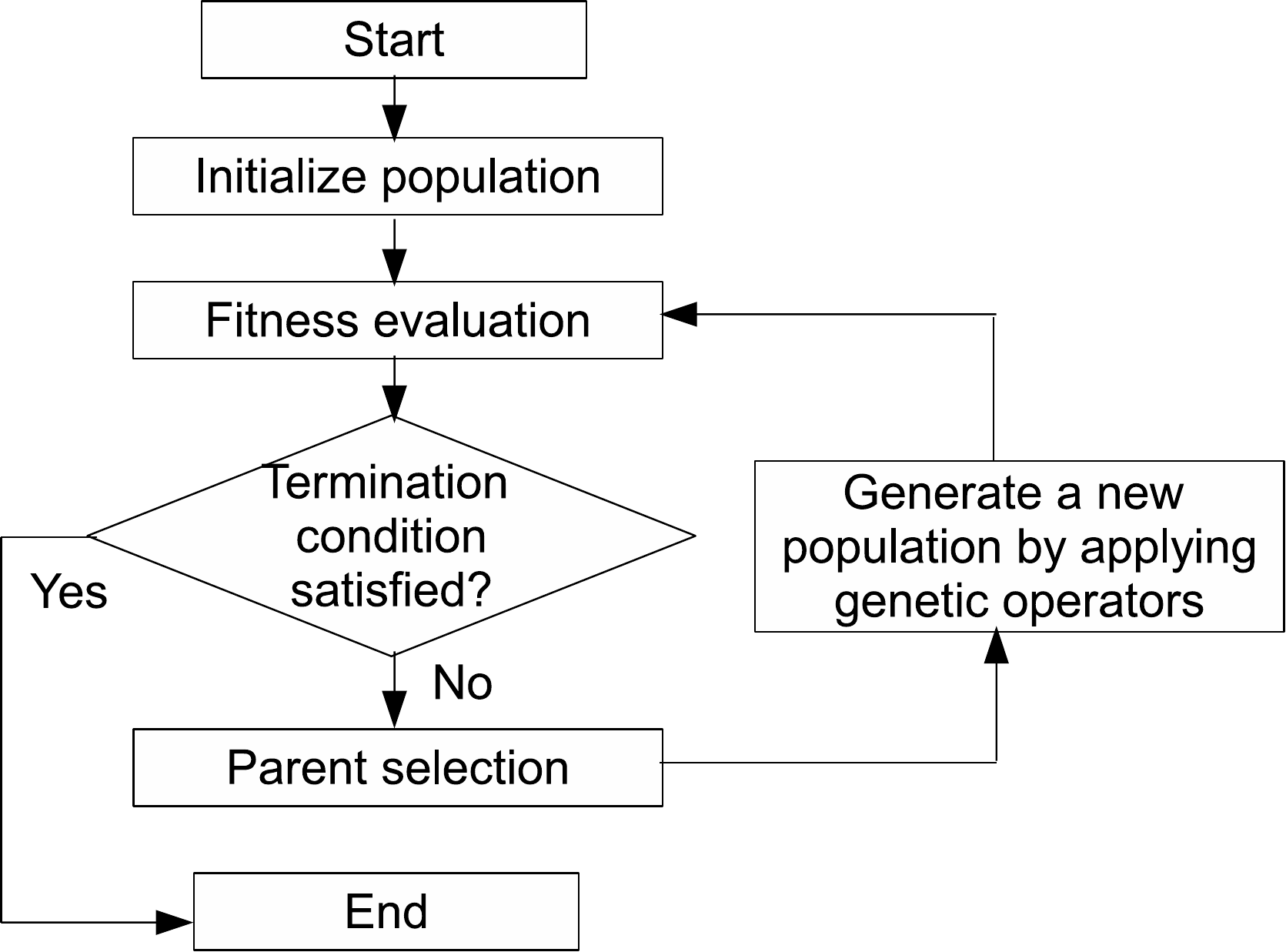}
\caption{Flowchart of Genetic Programming.}
\label{fig:Evolutionary Algorithm}
\end{center}
\end{figure}

The main steps of standard GP \cite{GPOTPOCBMONS} are as follows: 
\begin{enumerate}
\item random generation of an initial population of candidate solutions in tree form using the elements of function set and terminal set, selected by the modeler (user).
\item calculating fitness value of every individual of the population on the given training dataset (fitness cases)
\item selecting parents for mating based on the calculated fitness values, determined in previous step
\item applying sub-tree crossover and mutation (genetic operators) on selected parents for generating a new population of individuals.
\end{enumerate}

The process is repeated until the termination condition is fulfilled.

\section{Postfix-GP}
\label{sec:Postfix-GP}

\subsection{Features included in the Design of Postfix-GP}
The important features of the proposed Postfix-GP are categorized into:
(i) training dataset, function set, and terminal set related, (ii) GP parameters related, (iii) test dataset prediction related, (iv) GP run analysis related, and (v) serialization and de-serialization of GP experiments and results.

Training dataset, function set, and terminal set related features:
\begin{itemize}
\item Loading of training dataset
\item Loading of binary and unary functions
\item Loading of constants
\end{itemize}

GP parameters related:
\begin{itemize}
\item Selection of method to generate initial population
\item Configuration of GP parameters like population size, number of generations, crossover rate, mutation rate
\item Selection of crossover and mutation type
\item Selection of selection scheme
\end{itemize}

Test dataset prediction related:
\begin{itemize}
\item Loading out-of-sample test dataset for one-step predictions
\item Visualization of results of one-step predictions with statistical measures
\item Loading out-of-sample test dataset for multi-step predictions
\item Visualization of results of multi-step predictions with statistical measures
\end{itemize}

GP run analysis related:
\begin{itemize}
\item Plotting best adjusted fitness vs number of generations
\item Plotting average adjusted fitness vs number of generations
\item Plotting solution size vs number of generations
\end{itemize}

Serialization and de-serialization of GP experiments and results:
\begin{itemize}
\item Serialization of GP parameters, function set, terminal set, and obtained solutions to a file
\item De-serialization of GP parameters, function set, terminal set, and obtained solutions from a file
\end{itemize}

\subsection{Implementation of Postfix-GP}
\label{sec:Implementation}	
\begin{figure*}[!htbH]
		\centering
		\begin{tabular}{c}
		\includegraphics*[width=5.5in] {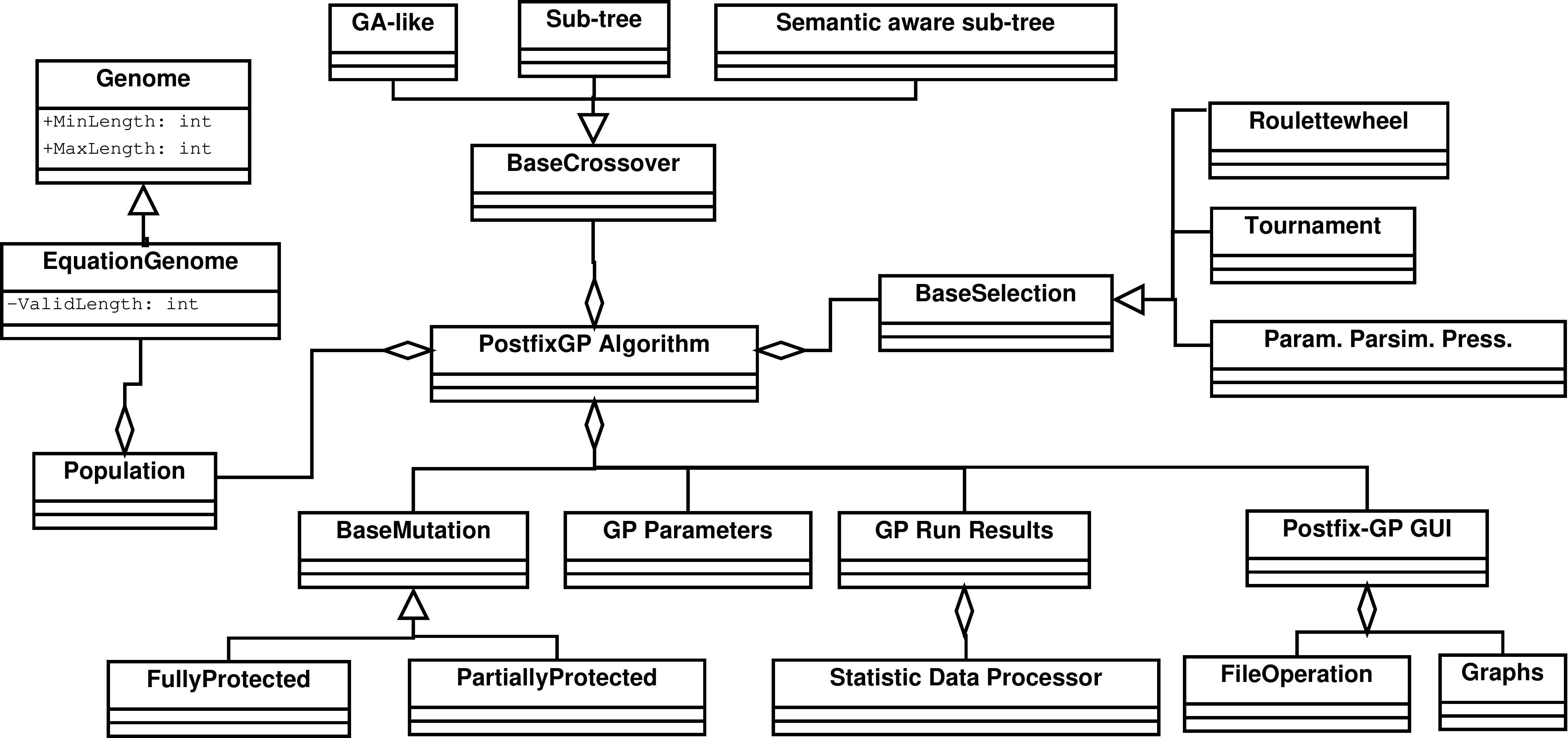}
		\end{tabular}
		\caption{Class diagram for Postfix-GP. }
		\label{fig:classdiagram}
		\end{figure*}

	This section presents the implementation details of Postfix-GP. This will be useful to the reader in understanding and customizing the proposed Postfix-GP framework. For details related to Postfix-GP solution representation scheme, refer \cite{dabhia2011}, \cite{dabhib2012}. Postfix-GP framework is developed using Microsoft .NET framework \cite{mdotnet} on Windows XP operating system. The ZedGraph \cite{zedgraph} class library is used for plotting the charts. ZedGraph is an open source graph library for .NET platform. \\
	The class diagram for Postfix-GP is depicted in \figurename~\ref{fig:classdiagram}. The classes can be grouped into following categories: (i) representation (\textsf{Genome}, \textsf{EquationGenome}), (ii) population (\textsf{Population}), (iii) crossover operator (\textsf{BaseCrossover}, \textsf{GA-like}, \textsf{Sub-tree}, \textsf{Semantic aware sub-tree}), (iv) mutation operator (\textsf{BaseMutation}, \textsf{Fully Protected}, \textsf{Partially Protected}), (v) selection schemes (\textsf{BaseSelection}, \textsf{Roulette-wheel}, \textsf{Tournament}, \textsf{Parsimony Pressure}), (vi) GP parameters (\textsf{GP Parameters}), and (vii) statistical analysis of results (\textsf{GP Run Results}).
	
\subsection{Individual, Population and Fitness}
	In Postfix-GP, an individual represents a candidate solution for a problem. We can create an instance of individual using \textsf{EquationGenome} class, which is derived from an abstract class called \textsf{Genome}. To generate an individual, we can use two-argument constructor, which returns a new individual having ValidLength in between MinLength and MaxLength.

	\textsf{
	public class EquationGenome() \{ \\
	\hspace*{5pt}public static int MinLength; \\
	\hspace*{5pt}public static int MaxLength; \\
	\hspace*{5pt}private int ValidLength;     \\
	\hspace*{5pt}private ArrayList TheArray = new ArrayList(); \\
	\hspace*{5pt}private int Rawfitness;\\
	\hspace*{5pt}private int Adjustedfitness;\\
	\hspace*{5pt}public EquationGenome(int Minlength, int Maxlength)\\
	\}
	}
	
	An individual comprises a genotype and a phenotype. The genotype is of type array of size equals to MaxLength specified by the user. In addition to the genetic material, an individual also contains raw fitness and adjusted fitness. The raw fitness is not range bound whereas the adjusted fitness can take values in the range of 0 to 1.\\
	In Postfix-GP, an instance of a population can be created using the class \textsf{Population}. An instance of \textsf{Population} contains a collection of instances of \textsf{EquationGenome}. Individuals can be added to and removed from the population (collection) using the iterator. \textsf{Iterator} is useful to iterate over a collection in sequential order. To create an instance of a population, several Postfix-GP parameters need to be passed. Following constructor can be used for the same.
	
	\hspace*{5pt}\textsf{
		public Population(CurrentDirectory, NoofGeneration,\\ NoofGenerationperCascade, PopulationSize, MinLength,\\
		MaxLength, MutationFrequency, CrossoverType, Cross-\\
		overFrequency, SelectionStrategy, Intervalarithmetic,\\ SemanticSensitivity, InitialPopulationType, MutationType)
		}
		
	The number of elites of a population can be set using \textsf{archivepopulationsize} parameter. By default, the archive size is set to $1/10$ of population size. The information related to an instance of a \textsf{Population} can be printed to a log file in human interpretable form. The logged population information is presented below. It begins by printing the number of generation, then details of individuals of current population and archive. For each individual it prints the genotype representation, its ValidLength and adjusted fitness value. \\
	\\
	\textsf{
		Generation 0 \\
		\hspace*{5pt}Population \\
		\hspace*{10pt} a\#-0.6\#a\#a\#*\#*\#-\#a\#-0.3\#0.8\#+\#-\#*\# $\to$ AdjFit $\to$ 0.0011 $\to$ ValPos $\to$ 13 \\
			\hspace*{10pt}a\#a\#-0.8\#-\#*\#a\#/\#-0.2\#a\#0.3\#-\#*\#+\#$\to$ AdjFit $\to$ 0.0010 $\to$ ValPos $\to$ 13 \\
			\hspace*{10pt}a\#-0.6\#a\#a\#*\#-0.8\#a\#a\#*\#*\#*\#+\#-\#-0.7\#+\# $\to$ AdjFit $\to$ 0.0002 $\to$ ValPos $\to$ 15 \\
			\hspace*{10pt}a\#a\#-0.4\#a\#*\#-\#a\#0.1\#S\#+\#*\# $\to$ AdjFit $\to$ 0.0003 $\to$ ValPos $\to$ 12 \\
			\vspace{5pt} \\
		\hspace*{5pt}Archive \\
			\hspace*{10pt}a\#a\#-0.4\#a\#*\#-\#a\#0.1\#S\#+\#*\# $\to$ AdjFit $\to$ 0.0003 $\to$ ValPos $\to$ 12 \\
			\hspace*{10pt}a\#-0.6\#a\#a\#*\#-0.8\#a\#a\#*\#*\#*\#+\#-\#-0.7\#+\# $\to$ AdjFit $\to$ 0.0002 $\to$ ValPos $\to$ 15
			}
	\\
		
Finally, a fitness value is assigned to an individual by evaluating it on training dataset. A code snippet for calculating a fitness of an individual is presented here. The code checks for type of terminal before performing any operation. \\ 
\\
\textsf{
public float PostFixCalculation(float[] inputs) \{ \\
		  \hspace*{3pt}equationstack.Clear(); \\
	       \hspace*{3pt}for (int i = 0; i $<$ this.ValidLength; i++) \{ \\
                    \hspace*{6pt}if ((int)TheArray[i] $<$ Operandcount) \{ \\
                        \hspace*{9pt}// Push a single input value on stack \\
                        \hspace*{9pt}equationstack.Push(inputs[(int)TheArray[i]]); \\
                    \hspace*{3pt}\} \\
                    \hspace*{3pt} else if ((int)TheArray[i] $<$ BinaryoperatorRange) \{ \\
                        \hspace*{6pt} if (equationstack.Count $>$ 0) \{ \\
                            \hspace*{9pt} y = (float)equationstack.Pop();  \\
                            \hspace*{9pt} if (equationstack.Count $>$ 0) \{  \\
                                \hspace*{12pt}x = (float)equationstack.Pop(); \\ 
                                \hspace*{12pt}result=DoOperation(x,y,Binaryoperators[(int)TheArray[i]$-$ \hspace*{76pt}OperandRange].ToString());  \\
                                \hspace*{76pt}if(double.IsPositiveInfinity(result) \\  \hspace*{80pt} $\Vert$double.IsNegativeInfinity(result)) \{ \\ 
	                                   \hspace*{103pt}return result;  \\
	                                \hspace*{96pt} \}  \\
	                            \hspace*{96pt}else  \\
	                                    \hspace*{100pt}equationstack.Push(result);  \\
                            \hspace*{9pt}\} \\
                       \hspace*{6pt}\} \\
                    \hspace*{3pt}\} \\
                    \hspace*{3pt} else if ((int)TheArray[i] $<$ UnaryoperatorRange) \{ \\
                    	................ \\
                    \hspace*{3pt}\} \\
			\}\\
		\} \\
	}
	
\subsection{Selection Methods}
	The task of selection mechanism is to select individuals from current population and previous (old) archive. Postfix-GP permits use of different selection mechanism for population and archive. All implemented classes for selection  mechanisms are derived from abstract class \textsf{BaseSelection}. The derived classes, \textsf{Roulettewheel}, \textsf{Tournament}, and \textsf{Parsimonypressure}, are required to implement the following two methods: \textsf{Select()} and \textsf{setIndividuals(ArrayList individuals)}. Each time the \textsf{Select()} method is called, it returns the index value. The individual at the given index value in the current population or archive is selected as one of the parents for crossover. The implementation of \textsf{Select()} method differs from one selection mechanism to another. \\
	\textsf{setIndividuals()}  method is used to fix whether the individuals are selected from the current population or from the old archive. \\
	\\
	\textsf{
	public int Select(); \\
    public void setIndividuals(ArrayList individuals);
    }
	\subsection{Genetic Operators}
			\subsubsection{Crossover}
			Postfix-GP provides implementation for following three types of crossover operators: (i) GA-like one-point crossover \cite{pcocofpgp}, (ii) sub-tree crossover \cite{pcocofpgp}, and (iii) semantic aware sub-tree crossover \cite{dabhib2012}, \cite{pcocofpgp}. The selection of crossover type is done by the user through GUI. The base class for all crossovers is \textsf{BaseCrossover}. The derived classes, \textsf{GAlike}, \textsf{Subtree}, and \textsf{Semantic aware subtree}, are required to implement the \textsf{Crossover()} method. The maximum number of trials parameter, used by semantic aware crossover, is by default set to value 20. \\
			\\
			\textsf{
			public abstract class BaseCrossOver \{ \\
		    \hspace*{5pt}    private int MaxCrossoverCount = 20; \\
		    \hspace*{5pt}    abstract public Genome[] Crossover(Genome gene1, Genome gene2); \\
			\}
			}

		\subsubsection{Mutation}
		Postfix-GP provides implementation for following two types of mutation operators: (i) fully protected and (ii) partially protected. The fully protected mutation selects an index value within the range $[1,ValidLength]$ and changes the element positioned at the selected index value to another element of the same arity. The partially protected mutation selects an index value within the range $[1,MaxLength]$. If the selected index value is less than MinLength then it applies fully protected mutation (an element positioned at selected index is replaced with another element of the same arity only) else it does not apply this constraint (an element positioned at selected index can be replaced with any other element having either same or different arity). The selection of mutation type is done by the user through GUI. The base class for all types of mutation is \textsf{BaseMutation}. The derived classes, \textsf{FullyProtected} and \textsf{PartiallyProtected}, are required to implement \textsf{Mutate()} method. The value of \textsf{MaxMutationCount} variable decides how many times Postfix-GP should try to ensure that the mutation operator generates a different gene than the original value. The default value for \textsf{MaxMutationCount} is set to value 10. The default value for mutation probability of operators (functions set) is set to value $0.6$. The code snippet for abstract class \textsf{BaseMutation} and \textsf{mutate()} method, implemented by \textsf{FullyProtected} class, are presented below. \\
		\\
		\textsf{
		public abstract class BaseMutation \{ \\
		\hspace*{5pt}  public int MaxMutationCount = 10; \\
		\hspace*{5pt}  private double operatormutatefrequency = 0.6; \\
		\hspace*{5pt}abstract public EquationGenome Mutate (Equation-\\
		\hspace*{20pt} Genome gene); \\
		    \}
		}
		
		\subsection {Sub-trees as Solutions}
		Postfix-GP derives all sub-trees of an individual having the Validlength greater than the MinLength and treats the extracted sub-trees as separate solutions (individuals) during the evolutionary process. This approach is useful to GP for exploring large solution search space and improving GP performance. Moreover, Postfix-GP uses elitism operator to preserve good solutions from one generation to the next generation without being affected by the destructive nature of crossover and mutation. Elitism also provides good building blocks for producing better solutions in successive generations. Postfix-GP preserves a set of highly fit solutions separately from the main population. Postfix-GP implements elitism using a fixed size archive. An archive is used to preserve the good solutions, found so far. 
		
		\subsection{Serialization/De-serialization of Population and GP-Parameters}
		Postfix-GP provides GUI for storing and retrieving of the population of individuals as well as GP-parameters from file system. The information about the final population and GP parameters are stored in binary format. The \textsf{ Serialize()} and \textsf{Deserialize()} methods of \textsf{ BinaryFormatter} class are used for storing and retrieving object information.
		
		\subsection{Statistics}
		
		Postfix-GP provides a facility to collect statistical details at every generation of a GP run. It stores data for individuals of both population and an archive into an excel file. It records following details for individuals of an archive per generation: (i) adjusted fitness value of the best individual of current generation, (ii) size of the best individual (solution), (iii) average adjusted fitness of archive, (iv) average node (element) count of archive, (v) best individual found so far, (vi) size of the best individual found so far, and (vii) Mean Absolute Error (MAE), Normalized Mean Square Error (NMSE), and Correlation Coefficient (CC) values for the best solution found so far. \textsf{WritePopulationAnalysisData()} and \textsf{WriteArchiveAnalysisData()} methods of \textsf{PostfixGPGUI} are used to store these data into the file.
		The mentioned statistical information are used to produce  the following graphs:
		\begin{itemize}
		\item Average adjusted fitness vs generations
		\item Average node count vs generations
		\item Best adjusted fitness vs generations
		\item Average adjusted fitness, average node count, and best adjusted fitness vs generations 
		\end{itemize}

		\subsection{Implementation of GUI}
					
		The class responsible for providing functionalities of GUI is named \textsf{PostfixGPGUI}. It is used to collect training dataset, function and terminal set, and Postfix-GP configuration parameters, provided by the user, and passing these data to Postfix-GP core. The GUI is useful to extend the functionalities of Postfix-GP core. The user can directly load the training dataset from stored comma separated (.csv) file. The \textsf{LoadDataFile()} method of \textsf{PostfixGPGUI} class is used to load the training dataset. The \textsf{LoadDataFile()} internally uses \textsf{getColumnNames(), getColumn()}, and \textsf{getRow()} methods of \textsf{FileOperation} class to determine the header of columns, number of columns, and number of rows in training dataset at run time. Then it calls \textsf{getData()} method of \textsf{FileOperation} class to load the training dataset into rows of \textsf{DataGridView}. 
				
		At the end of Postfix-GP run, GUI displays the best evolved solutions. The user can visualize the statistical details of any of the evolved solutions by pressing mouse button on the selected solution. The corresponding reference of an instance of \textsf{EquationGenome} is obtained from the current instance of \textsf{Population}. The \textsf{EquationGenome} reference is used to invoke \textsf{GetRawfitness(), GetAdjustedfitness(), GetCorrelationCoefficient(), CalculateNMSEFitness(), GetValidLength()} methods of \textsf{EquationGenome} class which returns the rawfitness, adjustedfitness, {\it r}, NMSE, and size of solution measures for the selected solution. Then, the instance of \textsf{Graphs} is created. The values for mentioned measures are passed as arguments of \textsf{DrawEquationMethod()} method of \textsf{Graphs} class. The \textsf{DrawEquation()} and \textsf{DrawGraph()} methods of \textsf{Graphs} class are useful to plot: (i) solution details and (ii) analysis of results of Postfix-GP run. 

\section{Case Study}
\label{sec:Case Study}
				
We have taken the benchmark symbolic regression problem, solved using GEP approach in \cite{PGEP}. Equation(1) is used to generate a set of twenty one equidistant points in range [-10:1:10]. These points are used as fitness cases (training dataset).
\begin{equation}
y = 3*(x+1)^3 + 2*(x+1)^2+ (x+1),
\label{Solution}
\end{equation}

The population size and the number of generations are set to values 50 and 200. The crossover and mutation rates are set to values 0.9 and 0.1.  The MinLength and MaxLength parameters of Postfix-GP are set to values 15 and 35. The terminal set T consists of the independent variable, x,  and a list of constants, T = \{ \begin{math} x \end{math} , list of constants \}. The selected constants values are \{1,2,3,5,7\}. The function set F consists of \{+, -, *, / \}.

\begin{figure}
\includegraphics[width=3.6in]{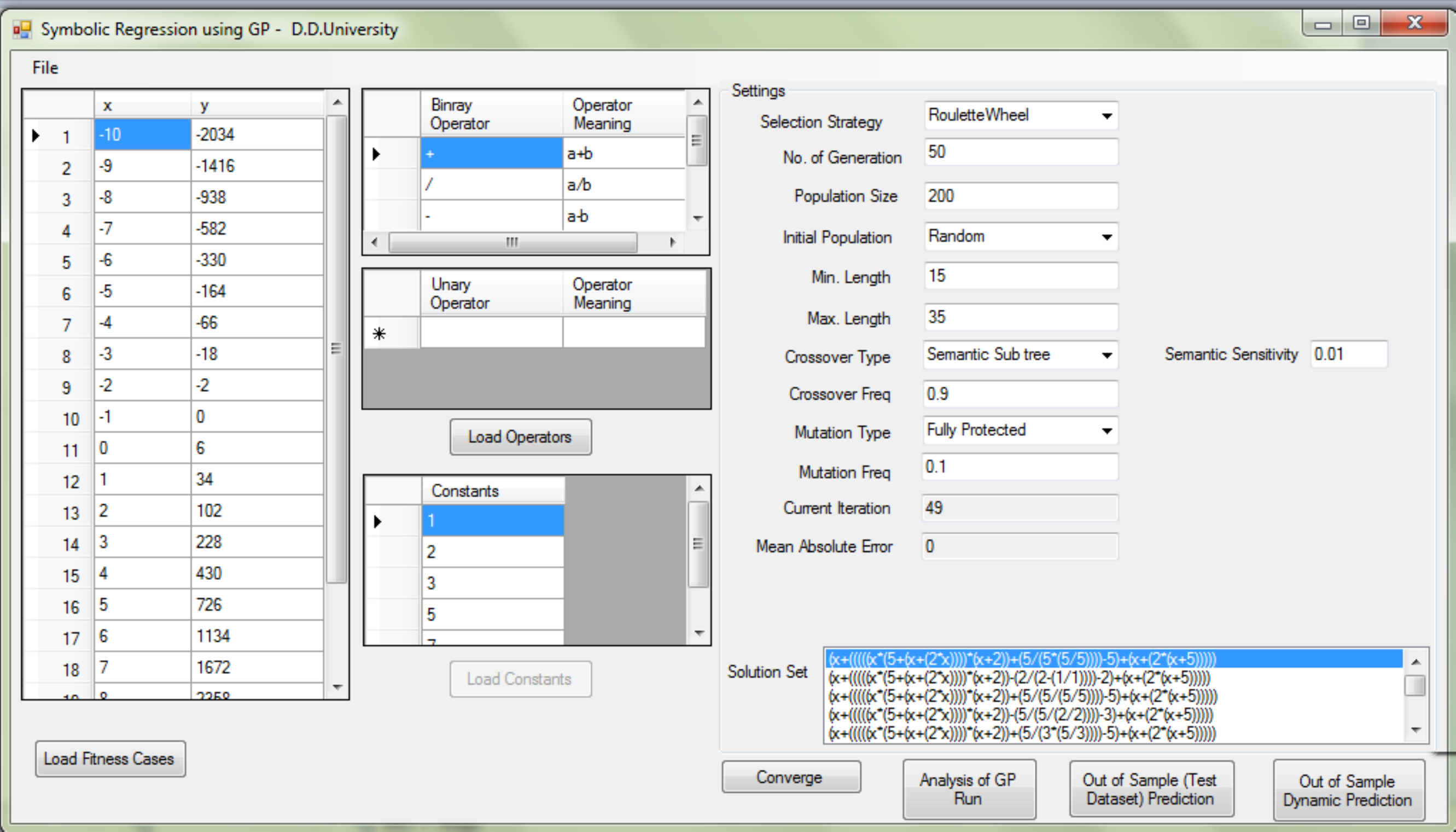}
\caption{Postfix-GP GUI for Experimentation}
\label{Fig:MainGUIPostfixGP}
\end{figure}

\begin{figure}
\includegraphics[width=3.3in]{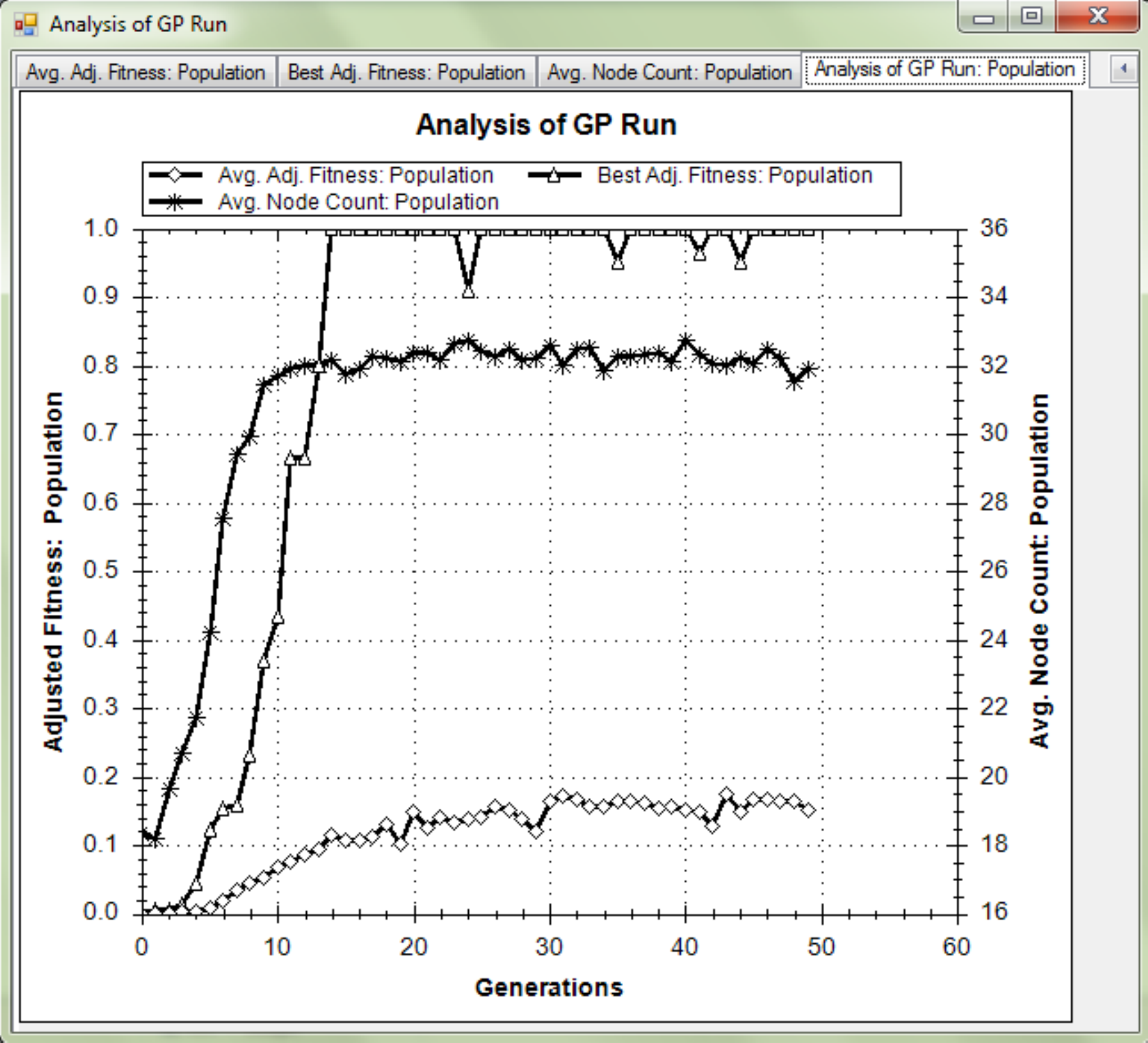}
\caption{Postfix-GP GUI for analysis of GP run}
\label{Fig:GPRunAnalysis}
\end{figure}

The left hand side panel of Postfix-GP GUI (\figurename~\ref{Fig:MainGUIPostfixGP}) provides the facility to load the training dataset, the function set, and the terminal set from the files. The accepted file format is comma separated values (.csv). The training dataset can be loaded from a file by activating {\it Load Fitness Cases} button of GUI. The function set and the terminal set can be loaded by activating {\it Load Operators} and {\it Load Constants} buttons of GUI. The right hand side panel of Postfix-GP GUI (\figurename~\ref{Fig:MainGUIPostfixGP}) provides facility to configure Postfix-GP parameters. \\
\begin{multline}
y = (x+(((((x*(5+(x+(2*x))))*(x+2))+(5/5* \\
(5/5))))-5)+(x+(2*(x+5)))))
\label{EvolvedSolution}
\end{multline}

\begin{figure}
\includegraphics[width=3.45in]{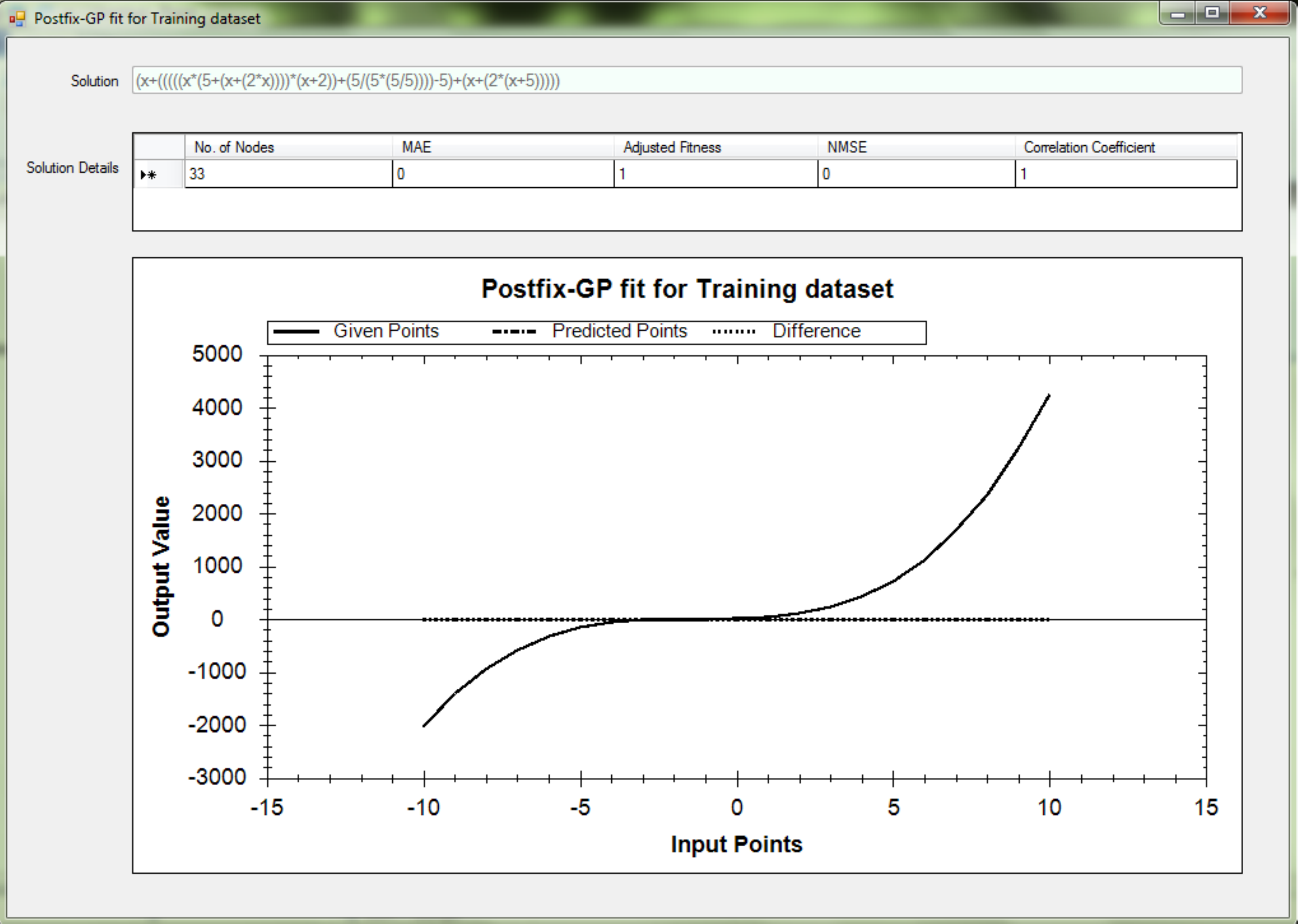}
\caption{Postfix-GP GUI showing statistical measures for the evolved solution for training data}
\label{Fig:Training}
\end{figure}

The separate GUIs are provided for visualization of: (i) evolved solution with its statistical measures, (ii) Postfix-GP run analysis, and (iii) Prediction of out-of-sample datapoints. \figurename~\ref{Fig:Training} presents visual representation of evolved solution with its statistical information. Equation ~(\ref{EvolvedSolution}) represents one of the best solutions found by Postfix-GP. Simplifying this equation results in equation ~(\ref{Solution}).

\figurename~\ref{Fig:Training} depicts the comparison of actual function and the function modeled by the evolved Postfix-GP solution. The evolved solution has structural complexity (number of nodes) of $33$. The statistical measures for the evolved solution for training dataset are as follows: $MAE = 0$, $NMSE = 0$, $Adjusted Fitness = 1$, and ${\it {r}} = 1$. 
\figurename~\ref{Fig:GPRunAnalysis} presents the visual representation of Postfix-GP run analysis over $50$ generations. It displays the plots of the average best fitness vs generation and the average solution size vs generation.

\begin{figure}
\includegraphics[width=3.45in]{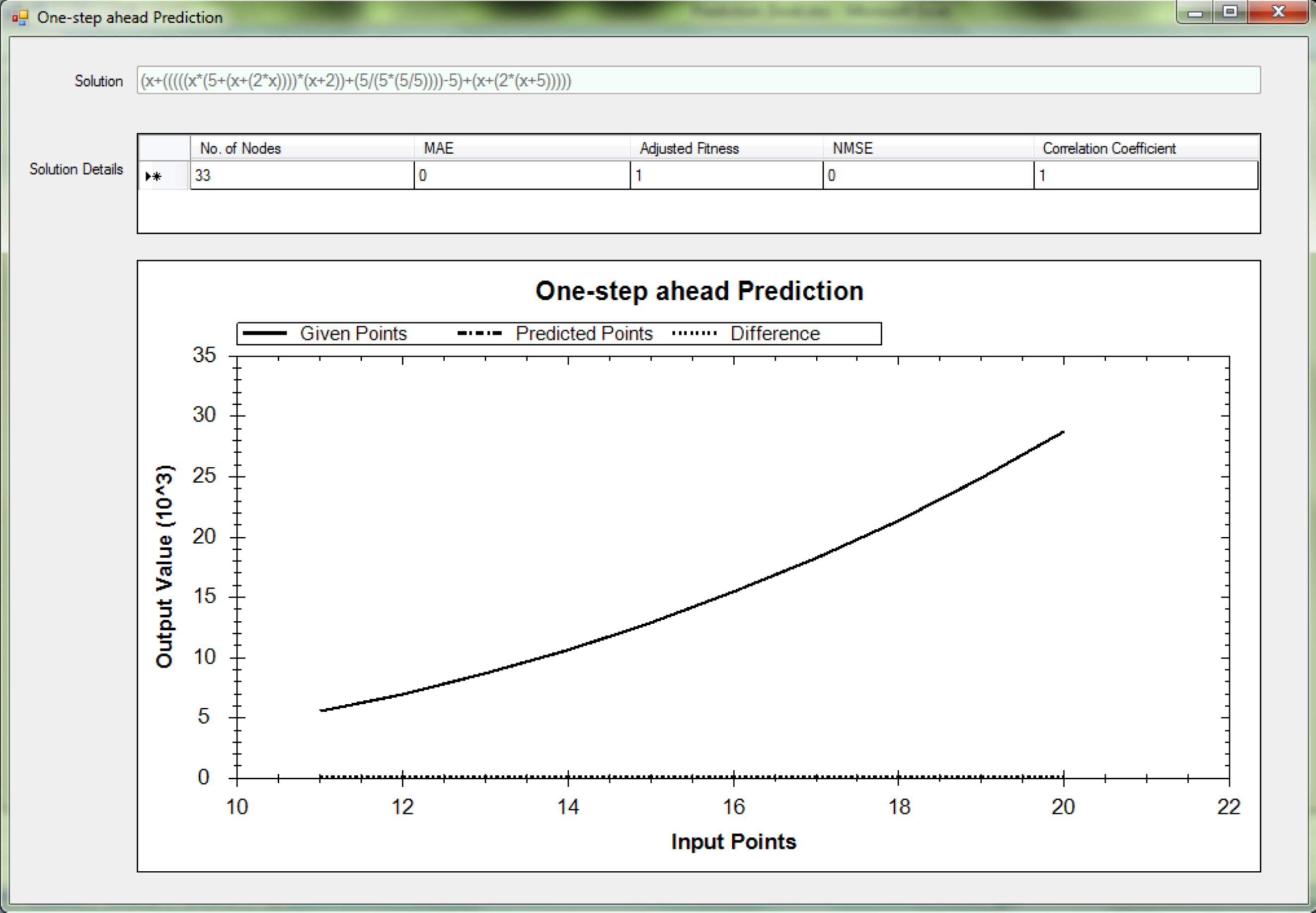}
\caption{Postfix-GP GUI for one-step prediction of out-of-sample data}
\label{Fig:OneStep}
\end{figure}

\figurename~\ref{Fig:OneStep} shows out-of-sample (test data) one-step ahead predictions obtained using the evolved solution for the values of \begin{math} x \end{math} in the range [11:1:20]. The statistical measures for one-step predictions are as follows: $MAE = 0$, $NMSE = 0$, $Adjusted Fitness = 1$, and ${\it {r}} = 1$. The error of zero value suggest that the one-step ahead predictions are very good.

\section{Features Comparison of Postfix-GP with lil-gp, ECJ, and JCLEC}
\label{sec:featurecomparison}
We have compared the features of Postfix-GP with lil-gp \cite{lilgp}, ECJ \cite{ecj}, and JCLEC \cite{jclec} GP systems. We have selected these GP systems because they are open source and freely available. Most of these systems do not provide easy to use functionality to end users. These systems can be used by persons having experience in the field of the evolutionary computing. Many systems are dedicated to a particular flavor of EC and others are generic (can be used for different flavors of EC). For example, lil-gp \cite{lilgp} provides support for GP only, whereas ECJ \cite{ecj} and JCLEC \cite{jclec} provide support for Genetic Algorithm (GA) \cite{AINAASAIAWATBCAAI}, Genetic Programming (GP) \cite{GPOTPOCBMONS}, and Gene Expression Programming (GEP) \cite{gepanaafsp2001}. However, for comparison of these systems with Postfix-GP, we have only considered GP flavor of these systems. \tablename~\ref{Tab: FeatureComparisonofPostfixGPlilGPECJJCLEC} presents the features comparison of Postfix-GP with these systems. \\

\begin{table*}
\tiny
\begin{center}
\caption {Comparison of the features of Postfix-GP system with other GP systems} \label {Tab: FeatureComparisonofPostfixGPlilGPECJJCLEC}
\begin{tabular}{p{0.2in}p{1.1in}p{1.1in}p{1.1in}p{1.0in}p{1.0in}}
\hline
{No.} & {Features} & {lil-gp} & {JCLEC} & {ECJ} & {Postfix-GP}  \\
\hline
	    1 & Genome representation & Tree  & Tree & Tree & Linear string \\
	
	    2 & GP parameters management & Text file (parameter file) & Configuration file & GUI configuration file & GUI \\
	
	    3 & Initialization technique & Full, Grow, Half-and-Half & Random, Koza & Random, Koza, PTC1, PTC2  & Random, Semantically diverse \\
	
	    4 & Crossover operators & Sub-tree & Sub-tree, Tree & Sub-tree & Sub-tree, GA-like one-Point, Semantic sub-tree \\ 
	    
	    5 & Limiting size of solution & Number of nodes/Depth limit & Number of nodes/Depth limit & Number of nodes/Depth limit & MinLength, MaxLength \\
	
	    6 & Selection schemes & Fitness proportional, Tournament, Random, Best, Worst & Elitist, Fitness proportional, Tournament, Random, MuLambda, Boltzman & Elitist,  Fitness proportional, Tournament, Random, MuLambda, Boltzman & Archive, Fitness proportional, Tournament \\
	
	    7 & Adding elements to function set & C function (function{.c,.h}) & Java file, Parameter file & Java file, Parameter file & C\# Function, Parameter file \\
	
	    8 & Cascading functionality & No & No & No & Yes \\

	    9 & Sub-trees as solutions & No & No & Yes (GEP) & Yes \\

	    10 & GP run statistics & Report files & GUI, Text file & .stat file & GUI, Text file \\

	    11 & GUI for prediction using evolved model & No & No & No & Yes\\
	     
	    12 & Generic & No & Yes & Yes & No \\ 
	    
	    13 & Learning curve & Small & Medium & High & Small \\ 
	
	    14 & GUI for visualization of statistical results & No & Yes & No & Yes \\
	    
	    15 & Architecture & Procedure oriented & Object oriented & Object oriented & Object oriented \\
	
	    
	    16 & Programming language & C & Java & Java & C\# \\
	\hline
\end{tabular}
\end{center}
\end{table*}
  
  \textbf{lil-gp}: lil-gp \cite{lilgp} is a GP toolkit implemented in C programming language. The toolkit is efficient and fast, as it is implemented in C. However,  it is difficult to extend the toolkit compare to other object-oriented implementations of GP systems. lil-gp uses only tree structure for an individual representation and provides limited fitness measures. Moreover, the toolkit does not provide graphical user interface to read an input (training) data file. The toolkit uses a parameter file to load GP parameters. The toolkit produces six reporting files (.sys,.gen,.prg,.bst,.his and .stt) that provide the statistical information of the GP run. There are many patches developed by different researchers to fix the bugs and to improve the functionality of the basic lil-gp.
  
  \textbf{ECJ}: ECJ \cite{ecj} is a Java based framework for evolutionary computation and genetic programming. ECJ is designed using the object-oriented concepts. Classes of ECJ framework are divided into four layers \cite{ecj}: (i) utility layer, (ii) basic and custom evolutionary computation layer, (iii) basic and custom genetic programming layer, and (iv) problem layer. As the framework is implemented in Java, it is slower in speed. Moreover, ECJ uses a tree (and not an arrays) to represent an individual, which requires dynamic memory allocation. Thus, the framework consumes large memory. ECJ determines GP parameters from a parameter file. ECJ determines classes to be loaded, the type of problem to be solved, the type of technique to use to solve the problem, and the way to report the statistical results of the run from the parameter file at the run time \cite{ecj}. This provides easy to extend functionality to the ECJ. ECJ stores statistical information of GP run in a text file. Moreover, it provides flexibility to produce the extra output files through class customization but does not provide a GUI to visualize this information.
  
  \textbf{JCLEC}: JCLEC \cite{jclec}  is a Java based framework for evolutionary computation and genetic programming. JCLEC \cite{jclec} is designed using the object-oriented concepts. The classes of JCLEC framework are divided in three layers: (i) system core,  (ii) experiments runner (reads an EA script file, execute all indicated algorithms and produce report files), and (iii) GenLab (a GUI on the top of experiments runner and system core layers, provides functionality to edit the experiment files and to view the GP run results) \cite{jclec}. GP parameters can be set by the user either through GenLab GUI or through an XML (configuration) file. However, the structure of configuration files is not user friendly. The framework provides the GUI to visualize statistical information of GP run. The JCLEC framework is easy to extend.

\section{Conclusions}
\label{sec:Conclusions}
This paper presented the design and implementation of Postfix-GP framework. The implementation details of Postfix-GP, including an individual representation, different crossover operators, mutations, and selection mechanism were also presented. Postfix-GP provides user interactive GUI for performing different activities. The user can load training dataset, function set, and constants. The user can set the GP parameters through GUI. The evolved solutions with their statistical measures can be visualized through GUI. Moreover, the user can also perform one-step and multi-step predictions using GUI. The evolved solutions can be stored in binary format and can be retrieved later on. The user can also analyze Postfix-GP run through GUI. Postfix-GP as a solution modeling tool was presented by solving symbolic regression problem. Postfix-GP addresses the requirements of ease of use and small learning curve before utilizing it to solve the problems. It was developed to minimize the user's time required to set up and run GP experiments.


\begin{thebibliography}{10}

\bibitem{lilgp}
D.~Zongker and B.~Punch, ``lilgp 1.01 genetic programming system,'' 1998.
  [Online]. Available: \url{http://garage.cse.msu.edu/software/lil-gp/}

\bibitem{ferreirac}
C.~Ferreira, \emph{Gene Expression Programming: Mathematical Modeling by an
  Artificial Intelligence}, 2nd~ed.\hskip 1em plus 0.5em minus 0.4em\relax
  Springer, May 2006.

\bibitem{silva}
S.~Silva and J.~Almeida, ``Gplab-a genetic programming toolbox for matlab,''
  pp. 273--278, 2003.

\bibitem{ecj}
S.~Luke, L.~Panait, G.~Balan, and Et, ``{ECJ 16: A Java-based Evolutionary
  Computation Research System},'' http://cs.gmu.edu/\~{}eclab/projects/ecj/,
  2007.

\bibitem{gagn}
C.~Gagn{\'e} and M.~Parizeau, ``Open {BEAGLE}: {A} new {C++} evolutionary
  computation framework,'' in \emph{GECCO 2002: Proceedings of the Genetic and
  Evolutionary Computation Conference}.\hskip 1em plus 0.5em minus 0.4em\relax
  New York: Morgan Kaufmann Publishers, 9-13 July 2002, p. 888. [Online].
  Available:
  \url{http://www.cs.ucl.ac.uk/staff/W.Langdon/ftp/papers/gecco2002/gecco-2002-15.pdf}

\bibitem{dabhia2011}
V.~Dabhi and S.~Vij, ``Empirical modeling using symbolic regression via postfix
  genetic programming,'' in \emph{Image Information Processing (ICIIP), 2011
  International Conference on}, 2011, pp. 1--6.

\bibitem{dabhib2012}
V.~Dabhi and S.~Chaudhary, ``Semantic sub-tree crossover operator for postfix
  genetic programming,'' in \emph{Proceedings of Seventh International
  Conference on Bio-Inspired Computing: Theories and Applications (BIC-TA
  2012)}, ser. Advances in Intelligent Systems and Computing, J.~C. Bansal,
  P.~K. Singh, K.~Deep, M.~Pant, and A.~K. Nagar, Eds., vol. 201.\hskip 1em
  plus 0.5em minus 0.4em\relax Springer India, 2013, pp. 391--402.

\bibitem{tsmapupgp}
------, ``Time series modeling and prediction using postfix genetic
  programming,'' in \emph{Advanced Computing Communication Technologies (ACCT),
  2014 Fourth International Conference on}, Feb 2014, pp. 307--314.

\bibitem{hwpmfrpoaroi}
V.~K. Dabhi and S.~Chaudhary, ``Hybrid wavelet-postfix-gp model for rainfall
  prediction of anand region of india,'' \emph{Advances in Artificial
  Intelligence}, vol. 2014, 2014.

\bibitem{jclec}
S.~Ventura, C.~Romero, A.~Zafra, J.~A. Delgado, and C.~HervÃ¡s, ``Jclec: a java
  framework for evolutionary computation,'' \emph{Soft Computing}, vol.~12,
  no.~4, pp. 381--392, 2008.

\bibitem{GPOTPOCBMONS}
J.~R. Koza, \emph{Genetic Programming: On the programming of computers by means
  of natural selection}.\hskip 1em plus 0.5em minus 0.4em\relax MIT press,
  1992, vol.~1.

\bibitem{mdotnet}
Microsoft, ``Microsoft .net framework software development kit,'' 2007.
  [Online]. Available: \url{http://msdn.microsoft.com}


\bibitem{zedgraph}
``Zedgraph,'' 2008. [Online]. Available:
  \url{http://www.sourceforge.net/projects/zedgraph}

\bibitem{pcocofpgp}
V.~K. Dabhi and S.~Chaudhary, ``Performance comparison of crossover operators
  for postfix genetic programming,'' \emph{International Journal of
  Metaheuristics}, vol.~3, no.~3, pp. 244--264, 2014.

\bibitem{PGEP}
X.~Li, C.~Zhou, W.~Xiao, and P.~C. Nelson, ``Prefix gene expression
  programming,'' in \emph{Late breaking paper at Genetic and Evolutionary
  Computation Conference {(GECCO'2005)}}, Washington, D.C., USA, 25-29 Jun.
  2005, pp. 25--31.

\bibitem{AINAASAIAWATBCAAI}
J.~H. Holland, \emph{Adaptation in Natural and Artificial Systems: An
  Introductory Analysis with Applications to Biology, Control and Artificial
  Intelligence}.\hskip 1em plus 0.5em minus 0.4em\relax Cambridge, USA: MIT
  Press, 1992.

\bibitem{gepanaafsp2001}
C.~Ferreira, ``Gene expression programming: a new adaptive algorithm for
  solving problems,'' \emph{Complex Systems}, vol.~13, no.~2, pp. 87--129,
  2001.
\end{thebibliography}
\end{document}